\pdfoutput=1

\documentclass[11pt]{article}

\usepackage{acl}

\usepackage{times}
\usepackage{latexsym}
\usepackage{inconsolata} 

\usepackage[T1]{fontenc}

\usepackage[utf8]{inputenc}

\usepackage{microtype}

\usepackage{graphicx}
\usepackage{tabularx}
\usepackage{booktabs}
\usepackage{multirow}
\usepackage{xspace}
\usepackage{array}
\usepackage{arydshln}

\definecolor{mdgreen}{rgb}{0.05,0.6,0.05}

\usepackage[normalem]{ulem}
\def\oddel#1{\bgroup\markoverwith{\textcolor{teal}{\rule[0.4ex]{2pt}{3pt}}}\ULon{#1}}
\def\opdel#1{\bgroup\markoverwith{\textcolor{red}{\rule[0.4ex]{2pt}{3pt}}}\ULon{#1}}
\def\psdel#1{\bgroup\markoverwith{\textcolor{violet}{\rule[0.4ex]{2pt}{3pt}}}\ULon{#1}}

\title{Three Ways of Using Large Language Models to Evaluate Chat}
\author{Ondřej Plátek, Vojtěch Hudeček, Patricia Schmidtová, Mateusz Lango \and Ondřej Dušek \\
Charles University, Faculty of Mathematics and Physics \\ Institute of Formal and Applied Linguistics \\ Prague, Czech Republic \\
\texttt{oplatek@ufal.mff.cuni.cz} \\
}

\begin{document}
\maketitle

\begin{abstract}
This paper describes the systems submitted by {\it team6} for ChatEval, the DSTC 11 Track 4 competition.
We present three different approaches to predicting turn-level qualities of chatbot responses based on large language models (LLMs).
We report improvement over the baseline using dynamic few-shot examples from a vector store for the prompts for ChatGPT.
We also analyze the performance of the other two approaches and report needed improvements for future work.
We developed the three systems over just two weeks, showing the potential of LLMs for this task.
An ablation study conducted after the challenge deadline shows that the new Llama~2 models are closing the performance gap between ChatGPT and open-source LLMs.
However, we find that the Llama~2 models do not benefit from few-shot examples in the same way as ChatGPT.
\end{abstract}

\section{Introduction}
\label{sec:intro}

This paper describes the systems submitted by {\it team6} for ChatEval, the DSTC 11 Track 4 competition aimed at evaluating open-domain chat.\footnote{
  Results \& task description at \href{https://chateval.org/dstc11}{chateval.org/dstc11}. Our experimental code is available at \href{https://github.com/oplatek/chateval-llm}{github.com/oplatek/chateval-llm}.}
We participated in Task 2, which focuses on evaluating multiple criteria on the level of individual dialogue turns.
The task of evaluating responses in a chat is challenging because it requires an understanding of the interlocutor's roles (pragmatics), the conversation's context, and the response's meaning (semantics).
At the same time, the conversations are often ungrammatical~\cite{rodriguezcantelar2023robust} and vary in style~\cite{zhang_personalizing_2018}.
The commonly used metrics, such as BLEU~\cite{papineni-etal-2002-bleu}, METEOR~\cite{banerjee2005meteor}, or BERTScore~\cite{zhang2019bertscore}, are based on comparison to human references and thus correlate poorly with human judgments on the turn-level, as they penalize many correct responses for a given chat context~\cite{zhao-etal-2017-learning}.
At the same time, human evaluation is expensive and time-consuming.
Previous referenceless metrics based on neural networks and language models still do not reach sufficient correlations with human judgements~\cite{deep-am-fm,lowe_towards_2017}.

\begin{figure}[t]
    \centering
    \includegraphics[width=0.4\textwidth]{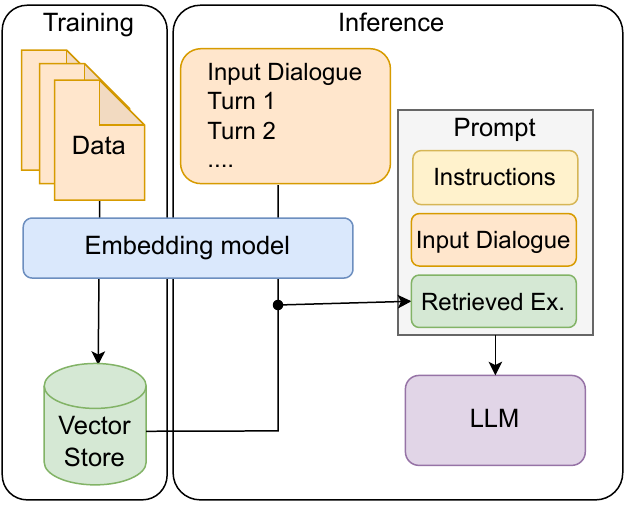}
    \caption{The architecture of the vector store approach with a LLM.
    During training, we construct the vector store from embedded annotated dialogues.
    At inference time, the input dialogue is embedded, and most similar examples from the vector store are retrieved to be included in the prompt.
    }
    \label{fig:vector-store}
\end{figure}

In our work, we followed up on the recent development of pretrained Large Language Models (LLMs) with instruction finetuning~\cite{brown_language_2020,raffel_exploring_2020}, which have been found to be capable evaluators in machine translation, summarization as well as dialogue~\cite{kocmi_large_2023,liu_g-eval_2023}.
Therefore, we applied LLMs and specific prompting to elicit ratings for the multiple qualities evaluated in DSTC11 Track 4 Task 2: appropriateness, content richness, grammatical correctness, and relevance.
We present three different systems used for our three submissions, all of which are based on LLMs and few-shot prompting:
(1) We evaluate a straightforward approach with manually designed fixed prompts for off-the-shelf open LLMs checkpoints.
(2) We train a simple feed-forward regression neural network (FNN) on top of frozen LLM embeddings to predict the turn-level metrics scores.
(3) We used the ChatGPT API and few-shot examples retrieved dynamically from the development set to improve the prompting performance.
As no data annotated with the target metrics were available for the challenge, we heuristically mapped existing annotations from the development set to the target metrics, and we manually annotated a small rehearsal dataset for hyperparameter search.

Based on the human annotations released after the challenge finished, our \textit{team6} achieved second place thanks to our third method,  
dynamically prompted chatGPT with few-shot examples.
This approach showed that LLM prompting is a viable option for prototyping chat evaluation.
However, the two other methods we explored scored worse: open LLMs with fixed prompts generally showed poor performance, and the regression FFN worked well on the development set but did not generalize well to the test set.

\section{Task \& Data}
\label{sec:task}

The goal of the DSTC11 Track 4 Task 2 was to predict several turn-level metrics automatically on the test set.
For each dialogue turn, considering the preceding dialogue history, the participants were to submit a system to predict the score of the target metrics, defined by the organizers as:
\begin{itemize}
    \item \emph{Appropriateness} -- The response is appropriate given the preceding dialogue.
    \item \emph{Content Richnes} -- The response is informative, with long sentences including multiple entities and conceptual or emotional words.
    \item \emph{Grammatical Correctness} -- Responses are free of grammatical and semantic errors.
    \item \emph{Relevance} -- Responses are on-topic with the immediate dialogue history.
\end{itemize}
Table~\ref{tab:rehearsal_examples} shows chat conversations from the rehearsal dataset with the turn-level metric annotations.

The organizers provided the participants with training, development, and test sets \cite{rodriguezcantelar2023robust}, each coming from different domains and annotated with different metrics:
\begin{itemize}
  \item \emph{Training set} -- consists of 390k dialogues, annotated with sentiment and toxicity labels. This set was not used in our experiments at all since our goal was to fine-tune or select LLMs that are already well-performing with no finetuning.
  \item \emph{Development set} -- consists of 24 datasets, some annotated with dataset-specific metrics. For our experiments, we created a \href{https://github.com/oplatek/chateval-llm/blob/807ebeeb812ab24df13d8cbb8fde24ac188bef7a/chateval/datasets.py#L354}{heuristic mapping} to the target metrics on a subset of the development set (see Section~\ref{sec:preprocessing}).
  \item \emph{Test set} -- consists of 3,470 dialogues and 130k turns, annotated with the target metrics. The data was only published in an anonymized form and at the end of the challenge, with no annotations or metadata, so that challenge participant could produce their model outputs. The annotations were published after the challenge was finished.
  \item \emph{Rehearsal set} -- this is a set of 156 turns collected in the same way as the test set, released earlier than the test set. 
We manually annotated this set with the target metrics (see Section~\ref{sec:preprocessing}) and used the result for hyperparameter search.
\end{itemize}

The submitted systems were benchmarked for the quality of their ranking using the Spearman correlation coefficients (SCC)~\cite{zar2005spearman} computed between the predicted scores and the human judgments.
As a secondary measure, the Pearson correlation coefficients (PCC)~\cite{freedman2007statistics} were used to evaluate the correlation.
The measures were computed for each of the target metrics separately.
The overall submissions' ranking was determined using the average of the four SCCs.

\begin{table*}[t]
\centering
\small
\begin{tabular}{lcccc}
\toprule
\textbf{Dialogue Turns} & \textbf{Appr} & \textbf{Rich.} & \textbf{Gram.} & \textbf{Rel} \\
\midrule
\small{My boss gave me a 10 raise just last month And it was a nice surprise} & 5 & 5 & 5 & - \\
\small{It's great and he might think you're doing a great job} & 5 & 5 & 5 & 5 \\
\small{We have always been very nice He has always been very supportive of me} & 4 & 5 & 5 & 5 \\
\small{That's a good thing} & 4 & 3 & 5 & 4 \\
\midrule
\small{do you have any pets?} & 5 & 4 & 3 & - \\
\small{I am retired so I love to travel so pets would slow me down} & 4 & 4 & 3 & 4 \\
\small{I understand that my idea of traveling is a hot hot bubble bath} & 3 & 4 & 2 & 2 \\
\small{Yes I have dogs and cats I like to take them with me on trips} & 2 & 4 & 2 & 2 \\
\bottomrule
\end{tabular}
  \caption{Two examples of complete conversations from the rehearsal set are annotated with turn-level metrics: appropriateness, content richness, grammatical correctness, and relevance.
  The context for each turn are the previous turns (lines) in the conversation.
  The second conversation at the bottom of the table shows an inappropriate response in the last turn because the last response contradicts previous responses of the system.}
\label{tab:rehearsal_examples}
\end{table*}

\section{Data Preprocessing}
\label{sec:preprocessing}

Since no information was provided on how the individual development dataset metrics relate to the target dialogue metrics, we built a heuristic to obtain target metric scores.
The heuristic uses a linear combination of one or more dataset-specific metrics to the target metrics,
chosen based on individual descriptions from the literature.\footnote{
  See the line 354 for the \href{https://github.com/oplatek/chateval-llm/blob/807ebeeb812ab24df13d8cbb8fde24ac188bef7a/chateval/datasets.py}{turn metric mapping} for different datasets.}
Using the development set and the heuristic, we created a supervised dataset and split it into training and development splits.
We used this {\it development dataset} for model selection or supervised training, and we use this dataset to develop the three systems described in~Section~\ref{sec:llms}.

\begin{figure}
  \includegraphics[width=\linewidth]{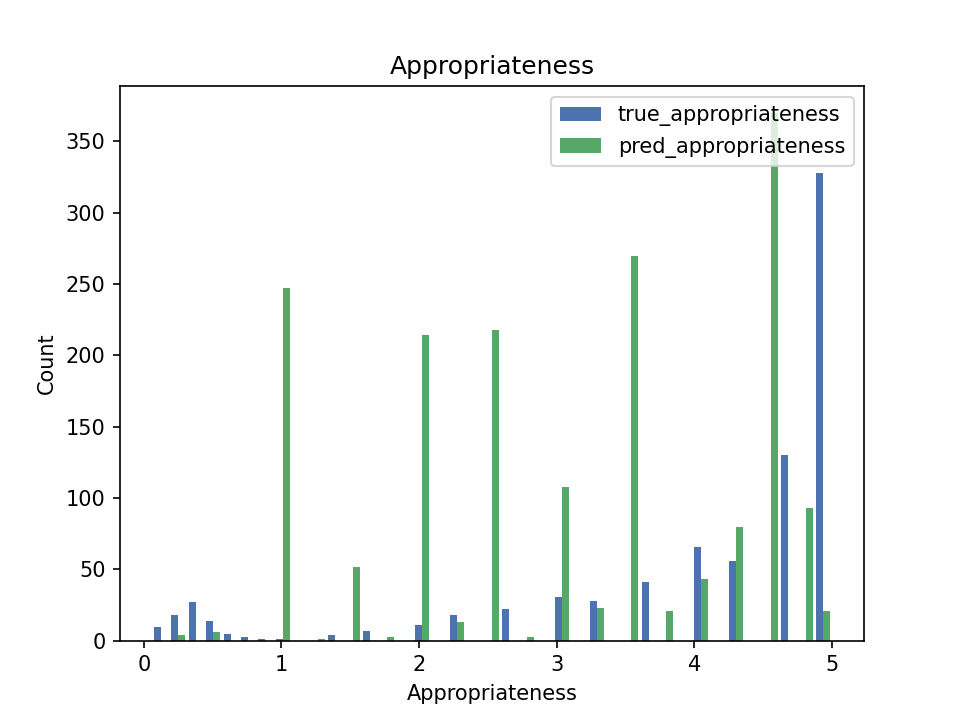}
  \includegraphics[width=\linewidth]{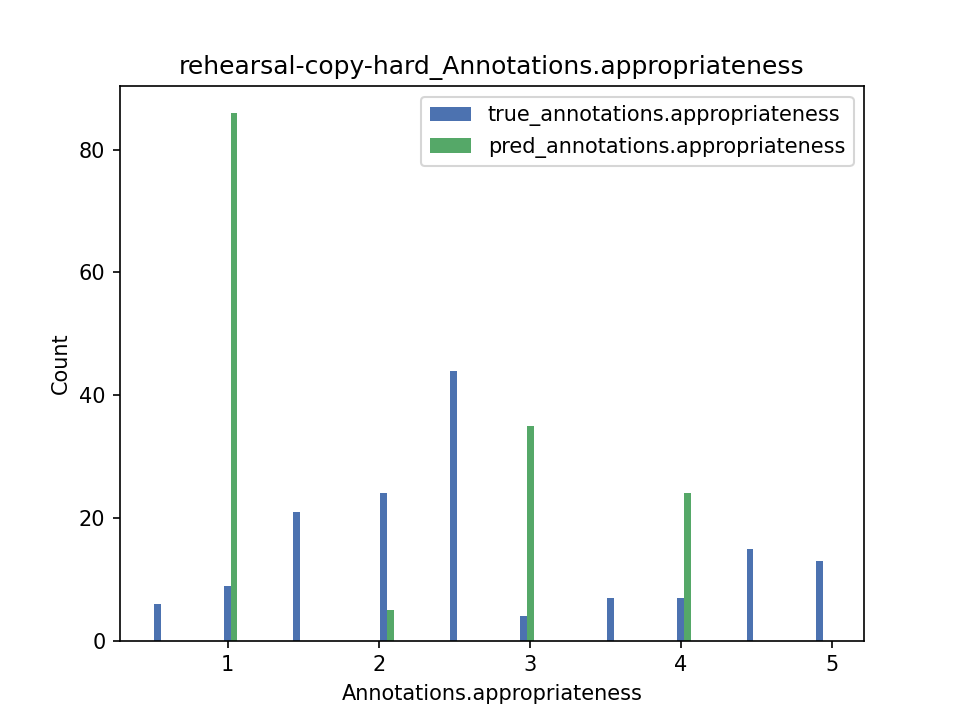}
  \caption{The histogram of predicted and human-annotated scores for {\it appropriateness} of a reply, on the test set (above) and on our manually annotated development set (below).
  Predicted scores are from ChatGPT with dynamic few-shot examples (see Section~\ref{sub:vector_store}).
  Note that the rehearsal set is not representative of the test set -- compare the blue bars representing the human-annotated scores.
  Interestingly, ChatGPT-predicted scores on the test set are not concentrated at the extremes, unlike on the rehearsal set.
  }
  \label{fig:appropriateness_histogram_test_rehearsal}
\end{figure}

During our experiments, we struggled to find representative labels and input data which could be used as a development set. Therefore, we decided to annotate the additional 156 turns from the {\it rehearsal set} with the target metrics described in Section~\ref{sec:task}.
We used this data to find our submitted systems' optimal hyperparameters.
We assumed that this data came from the same distribution as the test set, but this later proved clearly not to be the case, as seen in Figure~\ref{fig:appropriateness_histogram_test_rehearsal}.

Note that we did not use the training set at all.

\section{Submitted Systems}
\label{sec:llms}

Inspired by \cite{kocmi_large_2023}, we used pre-trained LLMs with prompts for predicting the individual metrics.
We started with the simplest approach possible and manually designed the prompts.

\subsection{Method 1: Simple Prompting}
\label{sub:simple_prompt}
We experimented with prompting GPT-NeoX-20B \citep{gpt_neox}, OPT-30B \citep{zhang2022opt}, and TK-Instruct-11B \citep{tkinstruct}.\footnote{The numbers identify each exact model checkpoint by the number of parameters.}
We tried several prompt templates for each model and selected the best-performing one on the development set and the manually annotated rehearsal set.
The templates were slightly adapted for each model to control for the deviations in model pretraining or instruction finetuning procedures, i.e., the wording of instructions or tags denoting a user-system interaction.

We used templates evaluating a single quality of each turn (i.e., calling the LLM four times to predict all metrics).
We focus on a single-metric template because most of the open-source models have trouble sticking to the desired output format when asked to generate a structured response with all four quality scores.
Our templates included two hardcoded examples from the DailyDialog set \cite{li_dailydialog_2017}, one of the provided development datasets.

We developed the prompt templates iteratively.
Every time we rephrased the prompt templates, we evaluated them on the DailyDialog dev set, which is part of the challenge dev set.

\subsection{Method 2: Feed-Forward Regressor on Top of LLMs}
\label{sub:regressor}
Our second method attempts to solve the problem that the prompted LLMs sometimes produce malformed output.
We assumed that LLMs extract relevant features even when the decoder produces a malformed one-best hypothesis.
Therefore, we aimed to use LLM contextual embeddings as features for a simple regressor.
However, instead of using the LLM's output directly, 
  we implemented a simple embedding extractor on top of the LLM, and we trained a regression model to predict all four scores based on the embeddings.
We use global max and average pooling over decoder layers and time steps of the decoded output to obtain the prompted response embedding.

\begin{figure}
    \centering
    \includegraphics[width=0.5\textwidth]{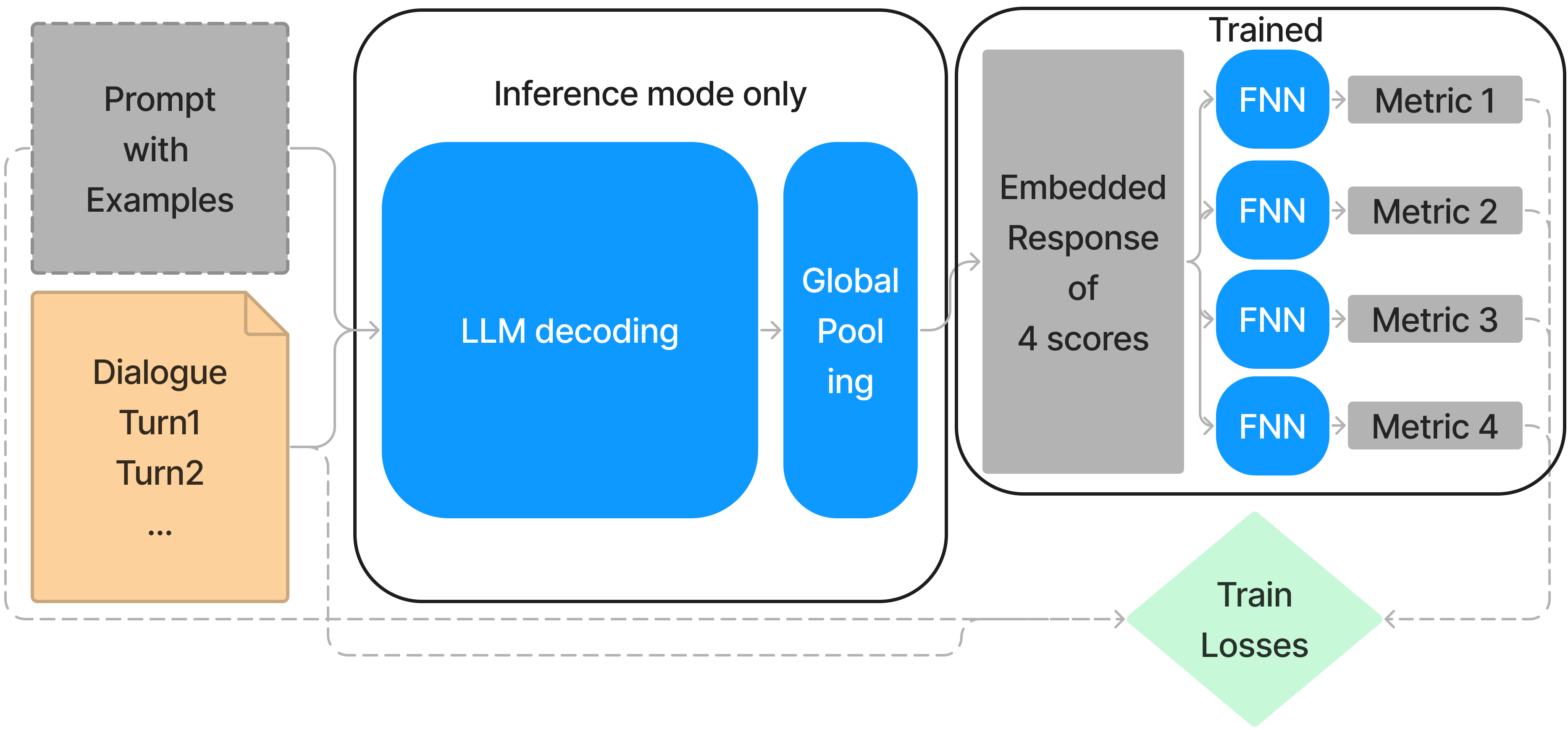}
    \caption{The architecture of the FFN trained on top of embeddings of LLM responses.}
    \label{fig:ffn}
\end{figure}

We designed the prompts so the LLMs' replies contain information about all four metrics, so a single call to the LLMs is sufficient to obtain all four scores.
At the same time, we designed the prompt so the LLM replies are as short as possible.
To train the regressor, we used our heuristically mapped development data (see Section~\ref{sec:preprocessing}).
We trained four simple feed-forward networks (FFNs), each modeling one of the target metrics using the same input embeddings. 
See Figure~\ref{fig:ffn} for the architecture of the FNN.

\subsection{Method 3: Dynamic Few-Shot Examples from a Vector Store}
\label{sub:vector_store}

The previous two approaches used fixed few-shot examples.
However, the performance of the in-context LLM learning can be improved by providing examples that are contextually similar to the instance being evaluated~\cite{brown_language_2020}.
We, therefore, implement a vector store with a dynamic few-shot example selection.
First, we take dialogues from the development set relevant to a given metric (based on our mapping described in Section~\ref{sec:preprocessing}), and compute turn-level embeddings.
These are then used as keys in a vector store optimized for similarity search.
At runtime, we retrieve a set of examples based on their similarity to the input and include them in the LLM prompt.
See Figure~\ref{fig:vector-store} for a detailed overview of the vector store architecture.

\section{Experiments}
\label{sec:experiments}

We experimented with the three methods described in Section~\ref{sec:llms}.
First, we we experimented with the Simple Prompting method using the open-source LLMs (Section~\ref{sub:simple_prompt}).
Based on the results, we started two independent experiments.
Section~\ref{sub:solution_ffn} describes the FFN training and Section~\ref{sub:solution_vector_store} describes the development of vector storage which we used with ChatGPT API.
For all three methods, we used the {\it rehearsal} set to select the best-performing model-template combination and hyperparameters.

\subsection{Simple Prompting Submision}
\label{sub:solution_simple}

For our baseline submission, we selected the best-performing model-template combinations for each quality separately and then combined the results.
Appropriateness and Relevance were generated by OPT-30b~\citep{zhang2022opt}.
Content Richness was generated by TK-Instruct~\citep{tkinstruct}.
As the outputs for ``Grammatical Correctness'' were malformed in most cases, we replaced the outputs with randomly generated scores.

\subsection{FFN Fine-Tuning setup}
\label{sub:solution_ffn}

We trained the FFN using two layers with 1024 hidden units and ReLU activation with batch size 2048 and learning rate 5e-5.
We used the log-cosh~\cite{saleh_statistical_2022} loss function.
We split the original development set into training and validation sets.
We trained until early stopping based on the validation set using SCC for {\it appropriateness} as a stopping criterion.
We extracted the embeddings from the prompted LLMs on the training and validation sets and cached them.
We used the same LLM checkpoints as in the simple prompting method.
We only used dev datasets whose annotations mapped to all four target metrics (see Section~\ref{sec:preprocessing}).;
  {\it DailyDialog}~\citep{li_dailydialog_2017}, {\it Fed-Turn}~\citep{mehri_unsupervised_2020}, {\it Persona-See}~\citep{see_what_2019}, and {\it Persona-Usr}~\citep{mehri_usr_2020}.

\subsection[short]{Vector Store Implementation}
\label{sub:solution_vector_store}
We use FAISS~\cite{johnson2019billion} to implement vector storage that can perform effective similarity-based retrieval.
To convert the dialogues into embeddings that are saved to the vector store, we used the MPNet~\cite{song_mpnet_2020} pretrained sentence representation model~\cite{reimers-2019-sentence-bert}.
We store the same development datasets in the vector store that we used for FFN training (Section~\ref{sub:solution_ffn}), with the heuristically mapped scores for all four metrics.

We used the prompt template in Figure~\ref{fig:template} with dynamically retrieved examples using vector store for the prompt and ChatGPT as the prompted LLM.\footnote{We used the \texttt{gpt-3.5-turbo-0301} API version.}
\begin{figure}
\begin{scriptsize}
\begin{verbatim}
Following is a dialogue context and the response to it.
Express how the response is appropriate given the context
with a continuous number between 1 and 5.
The higher the score, the more appropriate the sentences are.
Here are a few examples:
--------
{examples}
--------
Now complete the following with just a single float number:
Context: {dialogue_context}
Response: {response}
Appropriateness Score:
\end{verbatim}
\end{scriptsize}
\caption{
  Prompt template used with the few-shot dynamic examples retrieval with ChatGPT has a placeholder for the {\it examples}.
  Each {\it example} contains the turn {\it response} together with its {\it dialogue context} and the ground truth {\it appropriateness score}.
  The other methods used a similar template, with only a slight rewording.
} 
\label{fig:template}
\end{figure}

\section{Results \& Discussion}
\label{sec:results}
We report positive findings related to Method 3 (Section~\ref{sub:vector_store}), but we also report lessons learned from implementing the other two methods and, in general, using the data provided for the challenge.
First, we summarize observations from our use of the data (Section~\ref{sub:dataset_analysis}).
Then we report negative results from the simple prompting and FFN fine-tuning (Sections~\ref{sub:negative_simple} and~\ref{sub:negative_ffn}, respectively).
We also report our best results from the vector store (Section~\ref{sub:system_vs_turn}) and discuss what our best model in the challenge is capable of evaluating.
Finally, we add an ablation study in Section~\ref{sub:ablation_study} performed after the challenge was complete, comparing few-shot capabilities of ChatGPT with the newly released Llama~2 model.

We are aware that LLMs are trained on large datasets, some of which (e.g., ChatGPT) are not public.
However, due to the novelty of the test set~\cite{rodriguezcantelar2023robust}, we believe that the test set has not leaked to their training set.

\subsection{Dataset Analysis}
\label{sub:dataset_analysis}

The test set contains dialogue samples from various datasets unseen in the development and rehearsal sets: {\it BlenderBot3, ChatGPT, DSTC10Persona,  DSTC10Topical, ESL, GPT3, NCM}. 
The distribution of the test set was unknown to the participants, and most of the data comes from the unseen {\it BlenderBot3} and {\it ChatGPT} datasets.
We observed that scores for individual metrics were not normalized across the datasets as the {\it ESL} and {\it NCM} datasets had a range of 0-1, while the other datasets had a range of 1-5.

This discrepancy in data distributions most likely resulted in our model selection and hyperparameter search on the rehearsal dataset being detrimental to the final performance of our systems.
See the mismatch in the distribution of our own manual annotations on the rehearsal set and human annotation on the test set in Figure~\ref{fig:appropriateness_histogram_test_rehearsal}.
Furthermore, we argue that we could have achieved better results if we ran our model selection not only on the appropriateness metric but optimized for all four metrics.

\subsection{Simple Prompting is Fragile}
\label{sub:negative_simple}

In our informal experiments with simple prompting, we noticed that instruction-tuned LLM checkpoints produce results with intended formats more reliably.
We also experimented with templates evaluating all four metrics using a single prompt.
However, single-quality templates were generally more reliable and yielded outputs adhering to the expected formats more often.
We consistently observed that adding examples to the templates improved the reliability of the outputs.

Manual development of prompts, which relies on observing a small set of examples, was impractical for a diverse development dataset.
We frequently developed a promising prompt only to discover that the model produces malformed outputs when run on conversations from a different system.
The typical problem was that LLMs would interpret part of the input conversation as instruction. 
Consequently, instead of replying with the metric score, the model replied with a next turn fitting the conversation prefix.
Whenever the model did not respond in the desired format, we used an uninformed response score of 3.
The number of informed responses was the largest factor in the overall lower score for the simple prompting method.

\subsection{FFN is Fast but Lacks Normalization}
\label{sub:negative_ffn}

The training of the FFN is very efficient because we ran the LLMs only once in inference mode.
Note that the training was faster than extracting the embeddings from the LLMs, and a single FFN layer adds negligible computational and memory costs at inference time.
The FNN regression model solved the problem of LLMs producing malformed outputs.
However, our submission suffered from unnormalized scores in different development dataset splits, and the model performed poorly on the test set.
The results of our FFN training in Method 2
thus were influenced by incorrectly scaling the target metric values:
For example, the {\it FedTurn} scores lie in the range $[0,2.2]$ instead of $[1,5]$. 

\subsection{Are we Comparing Systems or Turns?}
\label{sub:system_vs_turn}

Method 3 (Section~\ref{sub:vector_store}) was the most successful in our experiments.
We argue that we could achieve even better results if we did model selection not only on the appropriateness metric but optimized for all four metrics.
We also argue that data mismatch between the rehearsal and test sets was detrimental to the performance of the systems.
Despite that, we placed second as a team, improved upon a baseline, and are relatively close to the best system in terms of the overall ranking.
See Table~\ref{tab:systems} for the comparison of the systems based on the average of the SCC over the four metrics.
See Figure~\ref{fig:appropriateness_histogram_test_rehearsal}.

\begin{table}[t]
\centering
\begin{small}
\begin{tabular}{llc}
\toprule
\multicolumn{2}{l}{\textbf{System }} &  \bf Avg.~Spearman \\
\midrule
\multicolumn{2}{l}{Baseline \cite{deep-am-fm}}       & 0.3387 \\
\midrule
\multicolumn{2}{l}{Winning submision (\emph{team4})} & 0.4890 \\
\midrule
\multirow{3}{*}{Ours:} & Simple Prompting  & 0.0807 \\
& FFN Regressor  & 0.1742 \\
& ChatGPT + Vector Store & 0.4190 \\
\bottomrule
\end{tabular}
\end{small}
  \caption{The overall performance of the baseline, the challenge winning submission and our three submissions.}
\label{tab:systems}
\end{table}

Our third method, ChatGPT with vector store examples (Section~\ref{sub:vector_store}), was the most successful in our experiments.
We observed that it easily contrasts between responses from different datasets but does not distinguish well among turns coming from the same dialogue system and the same dataset.
The SCC scores in Table~\ref{tab:spearman} shows that the score for the whole test set is better than most of the individual subsets based on different source datasets.


\begin{table}[h]
\centering
\begin{small}
\begin{tabular}{lccc}
\toprule
\textbf{Dataset} & Appr\tiny{opriateness} & Rel\tiny{evance} & Content \tiny{richness} \\
\midrule
\tiny{TEST-ALL}       & 0.488 & 0.361 & 0.452 \\
\midrule
\tiny{BLENDERBOT3}    & 0.383 & 0.287 & 0.303 \\
\tiny{CHATGPT}        & 0.122 & 0.060 & 0.181 \\
\tiny{DSTC10PERSONA}  & 0.803 & 0.968 & 0.216 \\
\tiny{DSTC10TOPICAL}  & 0.300 & 0.401 & 0.200 \\
\tiny{ESL}            & 0.199 & - & - \\
\tiny{GPT3}           & 0.091 & 0.007 & 0.242 \\
\tiny{NCM}            & 0.061 & - & - \\
\bottomrule
\end{tabular}
\end{small}
  \caption{The performance of our best system as Spearman correlation coefficients scores on the test set split for the metrics {\it Appropriateness}, {\it content richness}, and {\it relevance}.
  The first row {\it TEST-ALL} reports the results on the whole dataset.
  For brevity, we do not report {\it grammatical correctness} per splits which is 0.402 for the whole test set.
  The test set contains conversations from different systems, including ChatGPT and GPT3.}
\label{tab:spearman}
\end{table}

\begin{figure*}[t]
\centering
\includegraphics[width=0.45\textwidth]{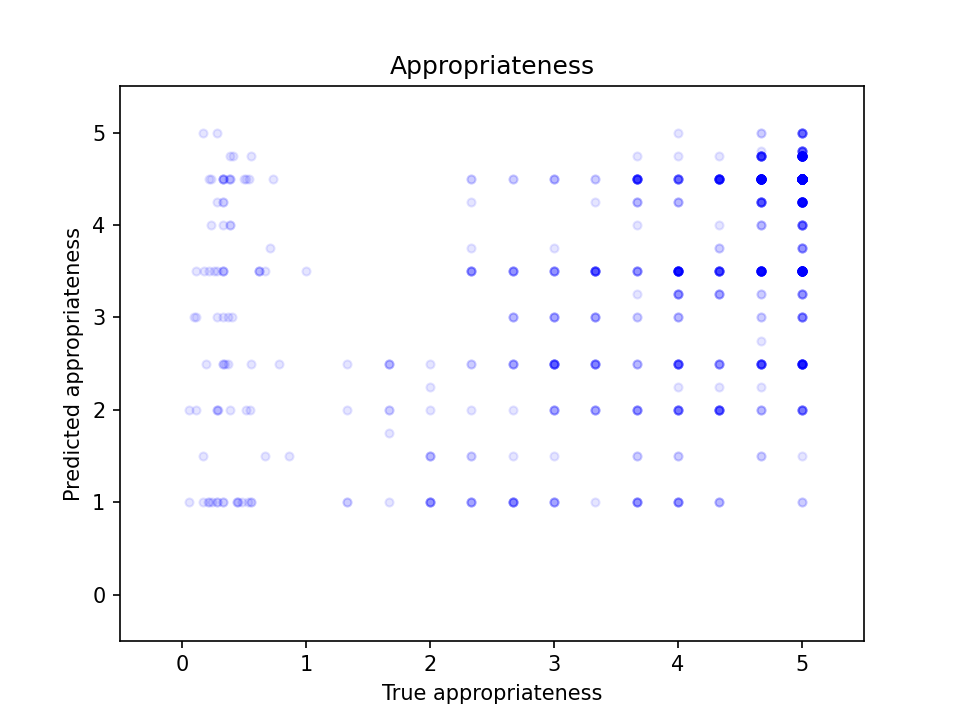}
\includegraphics[width=0.45\textwidth]{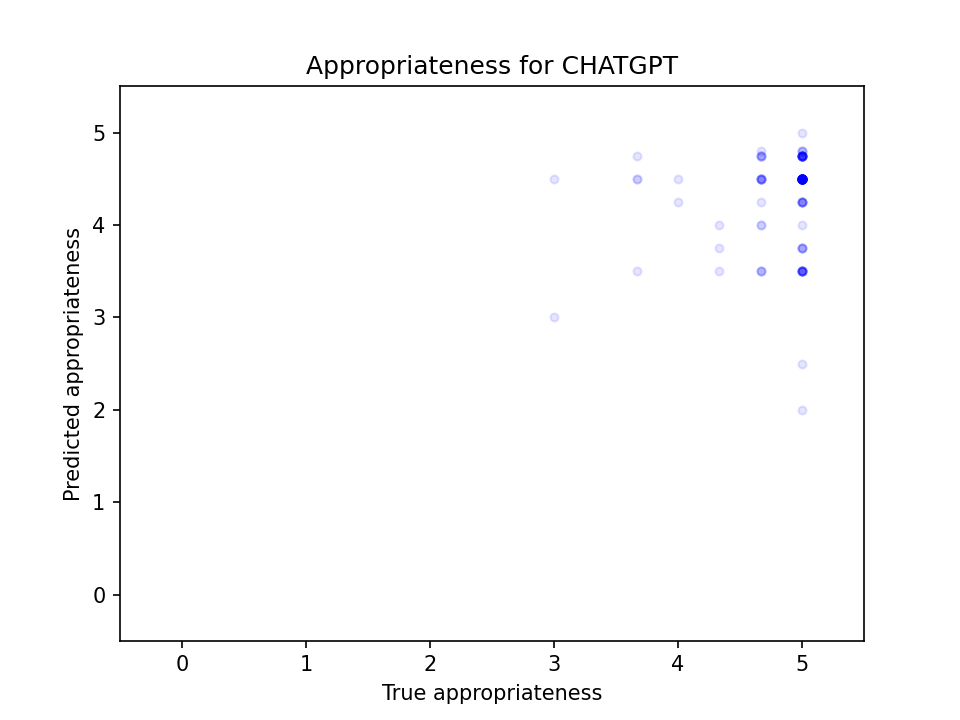}
\caption{
  The two scatterplots show the correlation of ground truth turn-level scores for appropriateness and the prediction of our best system, on the whole test set (left) and for the test set turns generated by ChatGPT (right).
Our system shows a relatively good correlation over the whole test set and 
 evaluates the ChatGPT results correctly as high-quality, but it fails at distinguishing the quality of individual ChatGPT turns.}
\label{fig:no_response_correlation}
\end{figure*}

\subsection{Revisiting Few-Shot Prompts in Ablation}
\label{sub:ablation_study}

\def\Porig{\emph{Porig}\xspace}
\def\Pnorm{\emph{Pnorm}\xspace}
\def\Pimpr{\emph{Pimpr}\xspace}

We present an additional ablation study, which we ran after the challenge was completed and evaluated on the \emph{Appropriateness} quality.
Using both ChatGPT and the newly released Llama~2 models \cite{touvron_llama_2023}, we investigate the influence of the few-shot examples on the performance of the models.\footnote{
  We used the Llama2-7b-chat-hf checkpoint (\url{https://huggingface.co/meta-llama/Llama-2-7b-chat-hf}) and the \texttt{gpt-3.5-turbo-0613} version of the ChatGPT API.
  The \texttt{gpt-3.5-turbo-0301} was used for the \Porig{} experiments with the original prompts from our submission.}
In order to do so, we made two changes to the prompts: (1) we designed a single prompt template that can be used both with and without few-shot examples,
(2) we normalized the use of newlines at the end of the prompt and in the few-shot examples, which improved performance.
We also (3) further improved the prompt by iterative experiments on the DailyDialog development set.

We label the improved prompt (with changes 1+2+3) as \Pimpr; we compare to a prompt closer to the original (with only changes 1+2 applied) as \Pnorm.
We then compared both ChatGPT and Llama~2 using both prompts \Pimpr{} and \Pnorm{} in three variants: (a) base without few-shot examples, (b) with two static examples (labeled \textit{-fix-2egs}), and (c) with two dynamically retrieved examples using the vector store (labeled \textit{dyn-2egs}, cf.~Section~\ref{sub:vector_store}.
We also include a comparison to the original ChatGPT with the prompt used in our model submitted to the challenge (labeled as \Porig, see Section~\ref{sub:solution_vector_store}). 
Finally, we ran an experiment with variants of \Porig/\Pnorm{} where we prompted the model to evaluate all the four qualities in a single prompt (labeled as \emph{-All}).


Our results in Table~\ref{tab:ablation} suggest that it pays off to design the prompt carefully, and it is beneficial to use few-shot examples in the prompts. However, using dynamic examples form the vector store instead of fixed ones does not bring further improvements.
We can see on ChatGPT results that our prompt improvements had an effect, and we were able to improve substantially over our challenge submission.
There is a notable gap between ChatGPT and Llama~2; on the other hand, the Llama~2 results are much better than any of our previous results with open models (see Sections~\ref{sub:negative_simple} and~\ref{sub:negative_ffn}).
We observe that predicting four qualities at once is not as good as predicting appropriateness only.
However, it still seems an attractive alternative since such template use is roughly four times more effective when predicting four qualities individually.
The percentage of failures for all reported systems is lower than 1\% and thus does not play a significant role in the evaluation.

\begin{table}[t]
\centering
\begin{small}
\begin{tabular}{l>{\hspace{-2mm}}lcc}
\toprule
\bf System & \bf Prompt &  \bf\hspace{-5mm} Spearman Appr.\hspace{-3mm} & \bf (\%fail) \\
\midrule
\multirow{4}{*}{\begin{tabular}{@{}l@{}}Llama 2\\\scriptsize 7B Chat\end{tabular}}
& \Pimpr & 0.3310 & (0.04\%) \\   
& \Pimpr-fix-2egs & 0.3756 & (0.56\%) \\  
& \Pimpr-dyn-2egs & 0.3683 & (0.36\%) \\ 
\midrule
\multirow{4}{*}{\begin{tabular}{@{}l@{}}ChatGPT\\\scriptsize 3.5-turbo-0613\end{tabular}} 
%
%
& \Pimpr & 0.4536 & (0.01\%) \\ 
& \Pimpr-fix-2egs & 0.6136 & (0.00\%) \\
& \Pimpr-dyn-2egs & 0.5962 & (0.00\%) \\
\midrule
\midrule
\multirow{4}{*}{\begin{tabular}{@{}l@{}}Llama 2\\\scriptsize 7B Chat\end{tabular}}
%
& \Pnorm & 0.3914 & (0.98\%)\\ 
& \Pnorm-fix-2egs & 0.3551& (0.06\%) \\ 
& \Pnorm-dyn-2egs & 0.3756 & (0.65\%) \\ 
& \Pnorm-All & 0.3710 & (0.01\%)  \\ 
\midrule
\multirow{2}{*}{\begin{tabular}{@{}l@{}}ChatGPT\\\scriptsize 3.5-turbo-0613\end{tabular}} 
& \Pnorm-dyn-2egs & 0.5462 \\
& \Pnorm-fix-All & 0.5334  \\
\midrule
\multirow{2}{*}{\begin{tabular}{@{}l@{}}ChatGPT\\\scriptsize 3.5-turbo-0301\end{tabular}} 
& \Porig-dyn-2egs & 0.4880 \\
& \Porig-fix-All & 0.3616 \\
\bottomrule
\end{tabular}
\end{small}
  \caption{Ablation study with the ChatGPT and Llama~2 7B Chat models for the \emph{Appropriateness} quality (see Section~\ref{sub:ablation_study} for prompt variants explanation).
  “\%fail” indicates the percentage of LLM outputs that failed to parse due to incorrect format.
  } 
\label{tab:ablation}
\end{table}

\section{Related Work}
\label{sec:related_work}

Recent works in chat evaluation focus on referenceless approaches, as these do not suffer from penalizing appropriate responses based on surface dissimilarity to a single human-written reference response \cite{liu_how_2017,lowe_towards_2017}.
Here, \citet{lowe_towards_2017} trained a neural network from scratch on relatively large annotated data to predict a single score, but this approach was later found to generalize poorly, even to basic data perturbations, let alone other datasets \cite{sai_re-evaluating_2019,lowe_introducing_2019}.

Later works leveraged pretrained language models for better generalization abilities,
such as BERT \citet{deep-am-fm,gao-etal-2020-dialogue}, RoBERTa \cite{mehri-eskenazi-2020-usr}, GPT-2 \cite{sinha-etal-2020-learning} or DialoGPT \citet{mehri_unsupervised_2020}.
These metrics are trained on human-labeled sets of system outputs based on popular open-domain datasets, similar to the ChatEval development data.
Some of them use additional data augmentation techniques, such as self-training \citet{mdd-eval}.
While they do achieve good correlations on some datasets, generalization with respect to unseen datasets is still not guaranteed \cite{yeh-etal-2021-comprehensive}.

\citet{sai2021perturbation} stressed the importance of predicting multiple qualities, such as, fluency and appropriateness, in dialogue evaluation.
At the same time, they asserted that metrics should be sensitive enough to distinguish between similar responses.
Using simple text perturbations targeting the individual qualities, they show that most existing metrics are not robust enough.

Two very recent works, closely related to ours, propose the usage of instruction-tuned LLMs to evaluate generated text in various tasks like summarization and dialogue response generation ~\cite{liu_g-eval_2023}, or machine translation~\cite{kocmi_large_2023}.
Both approaches use in-context learning and multiple prompting techniques to obtain scalar metric predictions or candidate rankings.
They achieved good results and correlations with human judgments.
However, they used only closed models for the evaluation and did not experiment with few-shot prompting using relevant examples.

\section{Conclusion}

We presented three simple approaches to using LLMs for turn-level chat evaluation.
We achieved promising results using ChatGPT prompting with few-shot example retrieval from a vector score, and ranked as the second-best team.
Based on the results of our best system, we argue that chat turn evaluation systems based on current state-of-the-art LLMs are usable only for system-level evaluation but not for segment-level evaluation,
i.e., they cannot distinguish between the quality of individual turns, especially for outputs of high-quality latest systems based on LLMs such as ChatGPT and GPT3.

We observed that LLMs are fragile to the prompts, few-shot examples and cannot be used out-of-the-box for chat evaluation.
We also report implementing a simple regressor on top of embeddings obtained from the prompted LLM decoder.
We attribute its poor performance to our incorrect implementation of data preparation.  

We also presented an ablation study that investigated the influence of the few-shot examples on the performance of LLMs.
We found that few-shot examples help the LLMs to generalize better to unseen data, especially with respect to fitting the desired output format.
However, using examples dynamically obtained from the vector store instead of hand-picked fixed examples did not bring any additional improvements. 

We reached a new best Spearman correlation coefficient of 0.6136 for appropriateness with ChatGPT and fixed few-shot examples in our ablation study.
In addition, the Llama~2 open model used in our ablation showed significant improvements over the challenge baseline.

\section{Acknowledgements}

This research was supported by Charles University projects GAUK 40222 and SVV 260575 and by the European Research Council (Grant agreement No.~101039303 NG-NLG). 
It used resources provided by the LINDAT/CLARIAH-CZ Research Infrastructure (Czech Ministry of Education, Youth, and Sports project No. LM2018101).
The authors thank the anonymous reviewers for their valuable feedback, Milan Fučík and Mateusz Krubiński for their suggestions and technical support.

\bibliography{semetric}
\bibliographystyle{acl_natbib}

\end{document}